\pdfoutput=1

\documentclass[11pt]{article}

\usepackage{emnlp2021}

\usepackage{times}
\usepackage{latexsym}

\usepackage[T1]{fontenc}
\usepackage{url}

\usepackage[utf8]{inputenc}

\usepackage{booktabs} 

\usepackage{microtype}

\usepackage{todonotes}
\usepackage{inputenc}

\title{CUNI systems for WMT21: Multilingual Low-Resource Translation for Indo-European Languages Shared Task}

\author{Josef Jon \and Michal Novák \and João Paulo Aires \and Dušan Variš \and Ondřej Bojar \\
       Charles University\\
  \texttt{\{jon,mnovak,aires,varis,bojar\}@ufal.mff.cuni.cz}  }

\begin{document}
\maketitle
\begin{abstract}
This paper describes Charles University submission for  Multilingual Low-Resource Translation for Indo-European Languages shared task at WMT21. We competed in translation from Catalan into Romanian, Italian and Occitan. Our systems are based on shared multilingual model. We show that using joint model for multiple similar language pairs improves upon translation quality in each pair. We also demonstrate that chararacter-level bilingual models are competitive for very similar language pairs (Catalan-Occitan) but less so for more distant pairs. We also describe our experiments with multi-task learning, where aside from a textual translation, the models are also trained to perform grapheme-to-phoneme conversion.
\end{abstract}

\section{Introduction}

The goal of the task was to translate text from Catalan into Occitan, Italian and Romanian. Additionally, use of parallel corpora which combine the evaluated languages with English, French, Portuguese and Spanish was permitted.  The choice of the languages from the same family invites to explore how to take advantage of similarities between the languages.

One way to exploit similarities between the languages translated by an NMT model is to train a single joint model for multiple languages. This way, parameters representing rules and features which are common for multiple languages can be shared and better estimated due to a larger amount of training examples related to them.

Another approach which can be effective when source and target languages are very similar is character-level processing of the text. Since most of the differences between Catalan and Occitan are straightforward orthographic variations, we hypothesize that the translation model would benefit from being able to manipulate the text at character level instead of larger subwords.

We also explore making use of language similarity in spoken form, aside from written form. 
Languages from the same language group may be more mutually intelligible in their spoken form rather than in the written form.
For instance, based on our anecdotal observations, native speakers of Czech report better understanding of spoken rather than written Polish.
This is mainly due to Polish orthography, which is regular but uses various digraphs, making Polish texts less comprehensible for common Czech speakers.
Phonemic representations may be even more helpful for languages with irregular spelling.

Instead of using automatically acquired phonemic representation as one of the inputs, we rather focus on strengthening robustness of our translation models by teaching them to produce this representation as an additional task.
Some of our models are thus trained to provide machine translation as well as grapheme-to-phoneme conversion (G2P) of the source.

\section{Main features of our approach}
\label{sec:mainfeats}

The core of our approach lies in leveraging multilingual training data, various subword granularity and phonemic representation of texts by multi-task learning.

All our models are instances of the Transformer architecture \cite{vaswani2017attention} as implemented in the MarianNMT \cite{junczys-dowmunt-etal-2018-marian}.
For the final submissions, we trained several models in multiple stages and tuned the decoding hyperparameters. Moreover, we applied character-level rescoring for the Catalan-Occitan submissions.

\subsection{Data preparation}
\label{sec:mainfeats-data}
In this section we describe our preprocessing steps, the relevant code is available at \url{https://github.com/ufal/bergamot.git/wmt21-multi-low-res}

\paragraph{Mulilinguality.}
 It has been shown (e.g. by \citet{zhang-etal-2020-improving}; \citet{Fan2020BeyondEM}; \citet{firat-etal-2016-zero};  \citet{tan-etal-2019-multilingual}; \citet{arivazhagan2019massively}; \citet{lakew-etal-2018-comparison}) that combining multiple translation directions into one model may be beneficial for the translation quality (especially for related languages) in the low-resource scenarios due to knowledge transfer between the translation directions, as it allows the model to get better estimates of the parameters that represent principles which are shared between the languages. 
 
 For our multilingual systems, we use the vanilla Transformer (single encoder, single decoder), concatenate the training data and insert a special token at the start of each source sentence to mark the desired target language, e.g. for translation from Catalan into Occitan:
 \texttt{<oc> Tres dels seus costats tenen porxada.}
 
 \paragraph{Subwords granularity and character-level translation.}
  It has been shown \cite{sennrich-zhang-2019-revisiting} that granularity of subword segmentation and thus the resulting vocabulary size has a large effect on translation quality in low-resource scenarios. For mid- and high-resource language pairs, vocabulary size of around 32k subwords is the usual choice. However, for smaller corpora, this size causes sparsity problems, since the vocabulary contains many subwords that were seen too few times to estimate sufficiently good embeddings for them. The solution is to split the words into smaller subwords or even into single characters. Moreover, we suspected that for similar languages, like Catalan and Occitan, small subword or character level translation may be beneficial because large part of the differences between the translations are merely orthographic variations and the ability to work on character level will allow the model to learn to perform these variations more easily.
 
\paragraph{Grapheme-to-phoneme conversion as an extra task.}
We hypothesize that teaching the model both to translate and to perform G2P may increase the model's robustness and consequently its performance.
Multi-task learning \citep{caruana97} has been successfully shown in NMT to either incorporate linguistic knowledge \citep{LuongLSVK15,eriguchi-etal-2017-learning,kiperwasser-ballesteros-2018-scheduled} or to exploit monolingual data \citep{wang-etal-2020-multi}.
Although it has been also used in G2P \citep{prabhu-kann-2020-frustratingly}, the two tasks has not been to the best of our knowledge modelled jointly so far.

Using a G2P tool, we prepare phonemic representation of the source side of the training data and combine it with the text data in two possible ways.

\textit{Vertical combination} is an analogy of how multiple translation directions are combined.
We concatenate the bitext with the data that consist of the same source side and its phonemic representation as the target side.
Furthermore, we use a special token at the start of each source sentence to indicate the G2P task, e.g. \texttt{<ca\_p>} for Catalan phonemization.

In \textit{horizontal combination}, we attempt to mimic multi-output learning \citep{multioutput19}, i.e. producing outputs for multiple tasks at the same time.
We thus enrich each target sentence with the phonemic representation of the source sentence.
The two are separated by a special symbol \texttt{<sep>}.
To evaluate the MT output, we need to strip off the phonemic part first.





\subsection{Model training and decoding}

\paragraph{Learning stages.}
Some of the models submitted to the shared task are a result of learning in two consecutive stages, each utilizing a different dataset.
In the pre-training stage, we build a general multilingual model, leveraging most of the available data sources.
In the fine-tuning stage, we continue training only on selected languages, possibly in conjunction with learning to convert graphemes to phonemes.

\paragraph{Decoding.}
During the beam search, we normalize the scores of each hypothesis by its length (the score is divided by $length^n$). We performed grid search over the $n$ coefficient and beams size for our primary submission and we obtained values $n=1.0$ and $b=8$. We used these values for all the systems.

\paragraph{Character-level rescoring.}
For Catalan-Occitan, we found character-level models to be competitive with subword models, but after manual inspection, we see some of the translations produced by these models included superfluous repetitions of groups of characters. For this reason, we decided to use the character-level model only for rescoring hypotheses produced by the subword-level models.

\section{Datasets}
\begin{table}[]

\resizebox{\columnwidth}{!}{
\begin{tabular}{ccccccc}
 & ca & en & fr &  it & oc & ro \\ \toprule
ca &- & 1305 & 2501 & 1756 & 57 & 1106 \\
en & - & - & - & 6434 & 37 & 1445 \\
fr & - & - & - & 21721 & 124 & 4815\\
\end{tabular}
}
\caption{Number of lines (in thousands) in corpora for each language pair used in our systems.}
\label{tab:datasets}

\end{table}

Apart from the Catalan, Occitan, Romanian and Italian data, we take advantage of the data in other languages allowed by the Shared Task organizers: Spanish, French and English (we did not use Portuguese corpora).
We used datasets specified by the task organizers, namely ParaCrawl, GlobalVoices, EuroParl, JW300, WikiMatrix, MultiCCaligned, Opus100, Books and Bible. Table \ref{tab:datasets} shows number of lines for each language pairs used in our experiments.

\section{Results}
In this section, we report BLEU \cite{Papineni02bleu:a} and ChrF2 cite{popovic-2015-chrf} scores on development and test sets provided by the organizers. We did not rerun test set evaluations for all the models, so for a small number of configurations we only show scores on the development sets.
\subsection{Tools}
We break the input text into subwords using SentencePiece~\cite{kudo-richardson-2018-sentencepiece}. We use MarianNMT \cite{junczys-dowmunt-etal-2018-marian} to train the models and the BLEU and ChrF scores are computed using SacreBLEU \cite{post-2018-call}. 
For experiments involving G2P conversion, we used \texttt{phonemizer} wrapper script\footnote{\url{https://github.com/bootphon/phonemizer}} around Espeak-ng speech synthesizer\footnote{\url{https://github.com/espeak-ng/espeak-ng}} to produce phonemic representation of the texts.

\subsection{Baselines}

\begin{table}[]
\small

\resizebox{\columnwidth}{!}{
\begin{tabular}{lcccccc}
\textbf{System}      & \multicolumn{3}{c}{\textbf{BLEU}}                                                                   & \multicolumn{3}{c}{\textbf{ChrF}}                                                                   \\ 
  & {\textbf{it}} &  \textbf{ro} & \textbf{oc} & \textbf{it} & \textbf{ro} & \textbf{oc} \\ \toprule
Opus-MT              & 32.4                            & -                               & 16.7                            & 0.608                           & -                               & 0.545                           \\
Google      & 32.3                            & 28.7                            & -                               & 0.609                           & 0.554                           & -                               \\
Apertium             & 32.1                            & 14.9                            & 67.0                            & 0.619                           & 0.461                           & 0.834                           \\
Bilingual   & 42.1                            & 29.8                            & 59.2                            & 0.674                           & 0.559                           & 0.789                          \\
Pivot    & 37.7 &	20.3 & 0.6 &0.636 &	0.505 &	0.082 

\end{tabular}
}
\caption{Results of the baseline system evalutation, development set.}
\label{tab:baselines_dev}

\end{table}

\begin{table}[]
\small
\resizebox{\columnwidth}{!}{
\begin{tabular}{lcccccc}

\textbf{System}      & \multicolumn{3}{c}{\textbf{BLEU}}                                             & \multicolumn{3}{c}{\textbf{ChrF}}                                             \\
\multicolumn{1}{l}{} & \textbf{it}          & \multicolumn{1}{c}{\textbf{ro}} & \textbf{oc}          & \textbf{it}          & \multicolumn{1}{c}{\textbf{ro}} & \textbf{oc}          \\ \toprule
Opus-MT              & 33.7                 &                    -             & 17.3                 & 0.612                &            -                     & 0.544                \\
Apertium             & 34                   & \multicolumn{1}{c}{13.3}        & 67.5                 & 0.624                & \multicolumn{1}{c}{0.408}       & 0.834                \\
Bilingual            & 44.9                 & \multicolumn{1}{c}{26.7}        & 59.4                 & 0.687                & \multicolumn{1}{c}{0.497}       & 0.787                \\
\multicolumn{1}{l}{} & \multicolumn{1}{l}{} &                                 & \multicolumn{1}{l}{} & \multicolumn{1}{l}{} &                                 & \multicolumn{1}{l}{}

\end{tabular}
}
\caption{Results of the baseline system evaluation, test set.}
\label{tab:baselines_test}

\end{table}

We used publicly available services and models as external baselines, and traditional bilingual Transformer models trained on provided corpora as our own baselines. We use SentencePiece preprocessing with 8k subword models for our bilingual baselines.
We also trained models to translate from Catalan to English and from English to the target languages to be able to do pivoted translation.
The external baselines include Google Translate (for Romanian and Italian), Romance multilingual model\footnote{\url{https://github.com/Helsinki-NLP/OPUS-MT-train/tree/master/models/ca+es+fr+ga+it+la+oc+pt_br+pt-ca+es+fr+ga+it+la+oc+pt_br+pt}} from Opus-MT project \cite{TiedemannThottingal:EAMT2020} and Apertium rule-based machine translation system \cite{Forcada2011ApertiumAF}, which was chosen since we suspected that the rule-based approach might work better than NMT for very low resource, but very similar language pairs, like Catalan-Occitan (and also Apertium is especially focused on  languages of that region).  Results on dev and test sets are presented in Tables   \ref{tab:baselines_dev} and  \ref{tab:baselines_test}, respectively.

We see that even our bilingual baselines outperform all other baselines aside from Apertium on Catalan-Occitan. We were unable to train functional English-Occitan model on the provided data (only 37k noisy sentence pairs), so the pivoted approach was not feasible in this direction.

\subsection{Improving bilingual models}
\begin{table}[]
\small

\resizebox{\columnwidth}{!}{
\begin{tabular}{lcccccc}
\textbf{BT}  & \multicolumn{3}{c}{\textbf{BLEU}}                                  & \multicolumn{3}{c}{\textbf{ChrF}}                                  \\
\multicolumn{1}{l}{} & \textbf{it}          & \textbf{ro}          & \textbf{oc}          & \textbf{it}          & \textbf{ro}          & \textbf{oc}          \\ \toprule
none          & 42.1 & 29.8 & 59.2                 & 0.674 & 0.559 & 0.789                \\
w, scr.    & 43.5 & 32.7 & 64.3                 & 0.680 & 0.584 & 0.818                \\
w, finet.   & -    & -    & 62.5                 & -     & -     & 0.810                \\
g, scr.    & -    & -    & 63.4                 & -     & -     & 0.815                \\
g, finet.   & -    & -    & 61.4                 & -     & -     & 0.803                \\
w(c)  & -    & -    & 64.7 & -     & -     & 0.819 \\
w(c) big  & -    & -    & 65.2                 & -     & -     & 0.821   
\end{tabular}
}
\caption{Adding backtranslation, development set. \textit{w} denotes backtranslated data originating from Wikipedia dumps, \textit{g} denotes general texts, \textit{scr.} denotes a system that was trained from scratch, \textit{finet.} denotes a system that was initialized by a baseline model trained on parallel data and finetuned, \textit{(c)} means character-level model and \textit{big} means that transfomer-big model was used instead of base.}
\label{tab:bt_dev}

\end{table}

\begin{table}[]
\small

\resizebox{\columnwidth}{!}{
\begin{tabular}{lcccccc}
\textbf{BT}  & \multicolumn{3}{c}{\textbf{BLEU}}                                  & \multicolumn{3}{c}{\textbf{ChrF}}                                  \\
\multicolumn{1}{l}{} & \textbf{it}          & \textbf{ro}          & \textbf{oc}          & \textbf{it}          & \textbf{ro}          & \textbf{oc}          \\ \toprule
none          & 44.9 & 26.7 & 59.4                 & 0.687 & 0.497 & 0.787                \\
w, scr.    & 45.8 & 28.4 & 64.3                 & 0.690 & 0.511 & 0.815                \\
w, finet.  & -    & -    & 62.4                 & -     & -     & 0.805                \\
g, scr.    & -    & -    & 63.6                 & -     & -     & 0.813                \\
g, finet.   & -    & -    & 61.6                 & -     & -     & 0.801                \\
w(c)  & -    & -    & 64.8 & -     & -     & 0.818 \\
w(c) big  & -    & -    & 65.2                 & -     & -     & 0.821              \\
\end{tabular}
}
\caption{Adding backtranslation, test set. Meaning of the rows is descirbed in previous table.}
\label{tab:bt_test}

\end{table}

\begin{table}[]
\small
\resizebox{\columnwidth}{!}{
\begin{tabular}{lcccccc}
\textbf{Vocab}       & \multicolumn{3}{c}{\textbf{BLEU}}       & \multicolumn{3}{c}{\textbf{ChrF}}       \\
\multicolumn{1}{l}{} & \textbf{it} & \textbf{ro} & \textbf{oc} & \textbf{it} & \textbf{ro} & \textbf{oc} \\ \toprule
8k           & 42.1 & 29.8 & 59.2 & 0.674 & 0.559 & 0.789 \\
2k           & 42.4 & 30.3 & 59   & 0.676 & 0.565 & 0.792 \\
char        & 38.8 & 28.6 & 62.6 & 0.652 & 0.555 & 0.808    \\
char-f & 41.2 & 28.3 & 62.1 & 0.669 & 0.554 & 0.808 \\

\end{tabular}
}
\caption{Results with varying vocabulary size, development set. \textit{Char-f} models are the original 8k models subsequently finetuned one character-level data.}
\label{tab:char_dev}

\end{table}

\begin{table}[]
\small
\resizebox{\columnwidth}{!}{
\begin{tabular}{lcccccc}
\textbf{Vocab}       & \multicolumn{3}{c}{\textbf{BLEU}}       & \multicolumn{3}{c}{\textbf{ChrF}}       \\
\multicolumn{1}{l}{} & \textbf{it} & \textbf{ro} & \textbf{oc} & \textbf{it} & \textbf{ro} & \textbf{oc} \\ \toprule
8k                  & 45   & 26.7 & 59.6 & 0.687 & 0.497 & 0.787 \\
2k                  & 44.5 & 26.1 & 59.1 & 0.685 & 0.495 & 0.788 \\
char               & 40.9 & 24.6 & 63.5 & 0.665 & 0.487 & 0.812 \\
char-f & 43.5 & 24.8 & 62.3 & 0.678 & 0.489 & 0.806 \\

\end{tabular}
}
\caption{Results with varying vocabulary size, test set. \textit{Char-f} models are the original 8k models subsequently finetuned one character-level data.}
\label{tab:char_test}

\end{table}
Before working on multilingual models, we focused on improving the bilingual systems to be sure our baselines are sufficiently strong. 

First, we add backtranslated data. We trained a joint multilingual model for translation from the target languages into Catalan. For Romanian and Italian, we used this model to translate Wikipedia,%
\footnote{We obtained the most recent dumps from \url{https://dumps.wikimedia.org/}}
for Occitan, we utilized Apertium and aside from Wikipedia, we also translated Occitan sides of all the other provided parallel corpora.  The results are presented in Tables \ref{tab:bt_dev} and \ref{tab:bt_test}. 
We see that backtranslation improves results for all the language pairs, and that for Occitan, wiki translation (rows marked as \textit{w}) works better than general corpora backtranslation obtained from Occitan sides of other parallel corpora (En-Oc, Fr-Oc and Es-Oc). We also observe that the performance is better when training with parallel and BT data from the beginning (\textit{scr.}), opposed to finetuning parallel-only trained model on parallel-BT mix (\textit{finet.}). 

We also tried to improve the results by choosing a correct subword granularity. We compared baseline models, which use SentencePiece vocabulary with 8k tokens, with 2k tokens and character level translation (see Tables \ref{tab:char_dev} and \ref{tab:char_test}).
Based on observations by \citet{libovicky-fraser-2020-towards}, we trained character level models both from scratch (row \textit{char}) and by finetuning the subword models (row \textit{char-f}).
We see that the character-level training works best for Catalan to Occitan translation.
We suppose it partially stems from the lack of resources for the language pair and partially from the relative similarity of the two languages. 

We combined the backtranslation and character level processing for Occitan to see if the improvements are orthogonal (Tables \ref{tab:bt_dev} and \ref{tab:bt_test} ). We also trained transformer-big models on the same data for comparison with larger models introduced in the next section.

\subsection{Multilingual models}

\begin{table*}[]
\centering
\resizebox{\linewidth}{!}{
\begin{tabular}{clcccccc}
   &                                    &             & \textbf{BLEU} &             &             & \textbf{ChrF} &             \\
\textbf{i}  & \textbf{Description}                        & \textbf{it} & \textbf{ro}   & \textbf{oc} & \textbf{it} & \textbf{ro}   & \textbf{oc} \\ \toprule
1  & ca-oc,ro,it                          & 43.7                 & 33.2                 & 63.8                 & 0.681                & 0.582                & 0.816                \\
2  & 1 + transformer-big                  & 43.1                 & 34.0                 & 63.5                 & 0.681                & 0.585                & 0.815                \\ \midrule
3  & ca,fr,es,en-oc,ro,it                 & 42.8                 & 33.7                 & 54.5                 & 0.675                & 0.584                & 0.761                \\
4  & 3 + balanced                         & 41.7                 & 33.6                 & 62.9                 & 0.667                & 0.583                & 0.806                \\
5  & 3 + balanced, bt                     & 41.8                 & 33.0                 & 60.3                 & 0.672                & 0.585                & 0.789                \\
6  & 3 + transformer-big                  & 44.7 &	35.1	&57.4 &	0.688	& 0.594 &	0.778 \\
7  & 3  +  transformer-bigger             & 42.6                 & 33.7                 & 52.1                 & 0.672                & 0.582                & 0.749                \\ \midrule
8  & 3 +  ca-es, ca-fr, ca-en             & 44.5                 & 34.6                 & 55.5                 & 0.686                & 0.591                & 0.769                \\
9  & 8 + big                              & 46.7                 & 37.1                 & 59.1                 & 0.700                & 0.607                & 0.792                \\
10 & 8 + bigger (430k updates)*           & $47.1^1$                 & $38.0^1$                  & 59.8                 & 0.702                & 0.613                & 0.794                \\
11 & 8 + bigger (2.1M updates, converged) & 48.5                 & 39.2                 & 62.7                 & 0.714                & 0.624                & 0.808                \\
12 & 10 + bt                              & $46.3^2$                  & $36.5^2$                  & 59.2                 & 0.701                & 0.608                & 0.792                \\
13 & 10 + finetuning for lang pair + bt   & 44.6                 & 34.4                 & 65.6                 & 0.689                & 0.597                & 0.824                \\
14 & 13 + char-level rescoring            & -                    & -                    & $67.1^1$                  & -                    & -                    & 0.833   \\            
\midrule
15 & 9 + ca-it,oc; vert. multi-task   & 45.2                    & -                    & 65.3                  & 0.690                    & -                    & 0.823               \\
16 & 9 + ca-it,oc; balanced vert. multi-task   & 42.9                    & -                    & 65.7                  & 0.675                    & -                    & 0.825 \\
17 & 16 + char-level rescoring   & -                    & -                    & $66.8^2$                  & -                    & -                    & 0.832
\end{tabular}
}
\caption{Results of our multilingual models, dev set. $^1$ marks our primary submissions, $^2$ is our secondary submission.}
\label{tab:multi_dev}

\end{table*}

\begin{table*}[]
\centering
\resizebox{\linewidth}{!}{
\begin{tabular}{clcccccc}
   &                                    &             & \textbf{BLEU} &             &             & \textbf{ChrF} &             \\
\textbf{i}  & \textbf{Description}                        & \textbf{it} & \textbf{ro}   & \textbf{oc} & \textbf{it} & \textbf{ro}   & \textbf{oc} \\ \toprule
1  & ca-oc,ro,it                        & 45.9        & 29.2          & 63.9        & 0.692       & 0.513         & 0.814       \\
2  & 1 + transformer-big                  & 45.7        & 29.0          & 63.2        & 0.691       & 0.511         & 0.808       \\ \midrule
3  & ca,fr,es,en-oc,ro,it               & 46.0        & 29.3          & 55.1        & 0.690       & 0.513         & 0.760       \\
4  & 3 + balanced                         & 45.0        & 29.1          & 63.3        & 0.684       & 0.511         & 0.803       \\
5  & 3 + balanced, bt                     & 44.3        & 28.9          & 60.8        & 0.685       & 0.515         & 0.788       \\
6  & 3 + transformer-big                  & 47.7        & 30.6          & 58.0        & 0.701       & 0.522         & 0.778       \\
7  & 3 + transformer-bigger               & 46.7        & 30.1          & 54.8        & 0.693       & 0.517         & 0.759       \\\midrule
8  & 3 + ca-es, ca-fr, ca-en             & 47.4        & 29.8          & 55.5        & 0.699       & 0.517         & 0.764       \\
9  & 8 + big                              & 49.1        & 31.7          & 59.5        & 0.710       & 0.531         & 0.788       \\
10 & 8 + bigger (430k updates)           & $50.5^1$        & $32.8^1$          & 60.3        & 0.717       & 0.533         & 0.792       \\
11 & 8 + bigger (2.1M updates, converged) & 51.1        & 33.9          & 62.6        & 0.722       & 0.544         & 0.804       \\
12 & 10 + bt                              & $49.5^2$        & $31.8^2$          & 59.9        & 0.713       & 0.533         & 0.792       \\
13 & 10 + finetuning for language pair + bt     & 47.3        &               & 66.6        & 0.702       &               & 0.825       \\
14 & 13 + char-level rescoring            & -           & -             & $66.9^1$        & -           & -             & 0.829    \\
\midrule
15 & 9 + ca-it,oc; vert. multi-task   & 48.6                    & -                    & 65.2                  & 0.706                    & -                    & 0.819               \\
16 & 9 + ca-it,oc; balanced vert. multi-task   & 45.3                    & -                    & 65.5                  & 0.687                    & -                    & 0.820 \\
17 & 16 + char-level rescoring   & -                    & -                    & $67.1^2$                  & -                    & -                    & 0.832
\end{tabular}
}
\caption{Multilingual models, test set. $^1$ marks our primary submissions, $^2$ is our secondary submission.}
\label{tab:multi_test}

\end{table*}
Our final submission is based on multilingual models. We combined the datasets allowed for the task and included a special language tag at the beginning of the source sentence to indicate the target language. The results on dev and test sets are presented in Tables \ref{tab:multi_dev} and  \ref{tab:multi_test}.  We use 32k vocabulary for the multilingual models.

Firstly, we trained a model only on the languages that were evaluated (system 1). We see that just by using the joint model, we obtained improved results for all language pairs. We also trained transformer-big model on the same data, as increasing model capacity usually improves performance especially for multilingual settings (system 2), but we observed same or worse results than with a base model.

Next, we added corpora with the other allowed translation directions which contain the evaluated languages on their target side, i.e. French, Spanish and English into Occitan, Romanian and Italian (system 3). At the first glance, including additional related languages did not improve the performance (and even hurts the performance for Catalan-Occitan), but we suspected that this might be a model capacity and data balancing problem. After oversampling the smaller training corpora to have the same number of sentences as the largest one, we see that performance of the model for this pair (4) reaches the levels of the previous model. Interestingly, adding backtranslated Wikipedia results in worse scores, even though backtranslation helped in bilingual models (5).
To see whether increasing the model capacity while using larger amount and more diverse training data is beneficial, we trained transformer-big (6) and transformer-big with 12-layer encoder instead of 6-layers, which we call transformer-bigger (7). For transformer-bigger, we used depth-scaled initialization proposed by \citet{zhang-etal-2019-improving}. We see that in fact, after adding more data, larger model capacity helps, but the 12-layered encoder transformer-big performs worse than the 6-layered one. We believe this is caused by instability of the training for the deeper models as in the next paragraph, we see improvements with the deeper model.

wUntil now, our goal was to mainly improve the target language generation by including other corpora with evaluated languages at the target side. We also tried to improve source-side Catalan encoding by adding corpora with Catalan on the source side, namely Catalan to French, English and Spanish (8). Resulting model shows improvements compared to the other language combinations, and again, increasing the model size ((9), (10) and (11\footnote{Model available at \url{http://hdl.handle.net/11234/1-3769}})) has even larger effect than for the previous models due to the amount and diversity of the training data. We hypothesize that increasing depth of the encoder helps in this case compared to the previous model because we added more data with Catalan source side and the increased encoder capacity could be used to learn more Catalan-specific features and rules.

Our primary submissions for Romanian and Italian are simply translations produced by the largest multilingual model (10). The training has not fully converged at the time of the submission and further training brought improvements in the range of 1-3 BLEU. Our secondary submissions for these two languages were the same models, however, we also included the backtranslated Wikipedia (12) in the training dataset.
Surprisingly, this approach lead to decrease in performance in terms of BLEU and ChrF2.
On the other hand, BERT and COMET scores in the official evaluation are same or slightly better for the models trained with backtranslation.

Due to the data imbalance, even the largest model underperforms in Catalan-Occitan. Because of the time constraints, we did not try oversampling Occitan corpora and training with balanced data, instead we fine-tuned the multilingual model for specific language pairs (13\footnote{Catalan-Occitan model available at \url{http://hdl.handle.net/11234/1-3770}}). Finally, we produced 20 best hypotheses for each sentence and rescored them by the character level Catalan-Occitan transformer-big introduced earlier (Table \ref{tab:bt_dev}), leading to a 1.5 BLEU increase on the dev set. This is our primary system for Catalan-Occitan.

Our submissions were ranked first in all directions with respect to all metrics except for the Catalan-Romanian BLEU score, where the M2M model was 0.2 points better (but after finishing the training, our model outperforms it by 0.8 BLEU).

For translation into Occitan and Italian, the organizers also performed human direct assessment evaluation. Translations produced by different systems were scored from 1 to 5 (on sentence-level, but document-level context was provided to the annotators). The results are shown in Table \ref{tab:human}.
\begin{table}[]

\tabcolsep=0.11cm \footnotesize	
\resizebox{\columnwidth}{!}{
\begin{tabular}{lllll}
\textbf{}                   & \multicolumn{2}{c}{\textbf{ca2it}}  & \multicolumn{2}{c}{\textbf{ca2oc}}  \\
                            & \textbf{z-score} & \textbf{raw}     & \textbf{z-score} & \textbf{raw}              \\ \toprule
HUMAN                       & 0.8$\pm$0.4          & 4.8$\pm$0.6          & 0.8$\pm$0.7          & 4.0$\pm$1.0  \\
\textbf{CUNI-Primary}       & \textbf{0.5$\pm$0.7} & \textbf{4.4$\pm$0.9} & \textbf{0.5$\pm$0.8} & \textbf{3.6$\pm$1.1} \\
M2M-100           & 0.4$\pm$0.7          & 4.2$\pm$1.0          & -0.7$\pm$0.8         & 2.0$\pm$1.0 \\
TenTrans-Primary            & 0.0$\pm$0.8          & 3.8$\pm$1.1          & 0.3$\pm$0.8          & 3.4$\pm$1.2 \\
BSC-Primary                 & -0.1$\pm$0.8         & 3.7$\pm$1.1          & 0.3$\pm$0.9          & 3.4$\pm$1.2 \\
UBCNLP-Primary              & -0.5$\pm$1.0         & 3.1$\pm$1.3          & 0.0$\pm$0.9          & 3.0$\pm$1.2 \\
mT5-devFinetuned & -1.2$\pm$0.9         & 2.3$\pm$1.2          & -1.0$\pm$0.7         & 1.7$\pm$0.9
\end{tabular}
}
\caption{Results of human evaluation performed by the organizers.}
\label{tab:human}
\end{table}

\subsection{Multi-task models}

In our experiments with multi-task learning, we trained the models to be able to both translate and perform G2P conversion of the source.
Using the \texttt{phonemizer} script, we automatically acquired phonemic representations of the Catalan sides in the Catalan-Italian, Catalan-Romanian and Catalan-Occitan data.
We then combined them with the original bitexts as proposed in Section~\ref{sec:mainfeats-data}.

As shown in Table~\ref{tab:phon}, we started with training multi-task transformer-base models from scratch using  vocabularies of 32k tokens.%
\footnote{%
Except for Occitan bilingual model, which uses a vocabulary of 8k tokens.
}
Apart from translation to Italian, multilingual models (in the bottom part) outperform the bilingual models (in the bottom).
In addition, vertical combination of texts and phonemes appears to perform better than the horizontal one.

Comparison of Tables~\ref{tab:phon} and~\ref{tab:multi_dev} suggests that even though trained from scratch multi-task learning seems to achieve competitive results for Catalan-Occitan.
We thus focus on this language pair in the following steps.
Interestingly, best scores for Occitan are achieved with a multilingual model that excludes Romanian.
We suppose Occitan is too distant from Romanian to benefit from it.
Therefore, we took the best-performing multilingual model at the time (system 9 in Tables~\ref{tab:multi_dev} and \ref{tab:multi_test}) and fine-tuned it with the Catalan-Italian and Catalan-Occitan training sets vertically combined with Catalan phonemes for these datasets (15).
As data balancing in multilingual models proved to be beneficial for Occitan, we also applied it before the fine-tuning, which results to even better performance for Occitan (16%
\footnote{%
Catalan-Occitan model available at \url{http://hdl.handle.net/11234/1-3772}}%
).
Finally, we rescored 20 best hypotheses by char-level Catalan-Occitan model as in the system 14, resulting in our contrastive submission for Catalan-Occitan (17).
Within all submitted Catalan-Occitan systems, our submission was ranked first in all metrics.


\begin{table}[t]
    \centering
    \resizebox{\columnwidth}{!}{
    \begin{tabular}{l ccc ccc}
           &   \multicolumn{3}{c}{\textbf{BLEU}}       &  \multicolumn{3}{c}{\textbf{ChrF}}  \\
    \textbf{Description} & \textbf{it} & \textbf{ro} & \textbf{oc} & \textbf{it} & \textbf{ro} & \textbf{oc} \\
    \midrule
tgt horiz. & 43.2 & 31.0 & 62.5 & 0.680 & 0.568 & 0.811 \\
tgt vert. & 43.2 & 31.6 & 63.6 & 0.679 & 0.573 & 0.817 \\
\midrule
it,oc horiz. & 43.3 & -- & 63.9 & 0.681 & -- & 0.818 \\
it,oc vert. & 42.9 & -- & 64.5 & 0.678 & -- & 0.821 \\
it,ro,oc horiz. & 42.5 & 32.8 & 63.4 & 0.675 & 0.578 & 0.814 \\
it,ro,oc vert. & 43.1 & 32.8 & 63.4 & 0.678 & 0.579 & 0.815 \\
    \end{tabular}
    }
    \caption{Results of multi-task models on dev set. The source side always consists of Catalan texts. The top part shows bilingual models, while the models in the bottom part are multilingual.}
    \label{tab:phon}
\end{table}

\section{Conclusion}
We described our submission to the shared task, which ranked first according to the majority of the used metrics for all languages. We used multilingual transformer models and we present results showing that combining all the languages into  single model improves upon bilingual baseline by a large margin. We also present our findings about using multi-task learning, where aside from translation of the source, the model also learns to convert the source sentence from graphemes to its phonemic form.

\section*{Acknowledgements}
Our work is supported by the grants H2020-ICT-2018-2-825303 (Bergamot) of the European Union,
19-26934X (NEUREM3) of the Czech Science Foundation
and SVV 260 575.
Our work has also been using data provided by the LINDAT/CLARIAH-CZ Research Infrastructure, supported by the Ministry of Education, Youth and Sports of the Czech Republic (Project No. LM2018101).
\bibliography{custom}
\bibliographystyle{acl_natbib}

\end{document}